\documentclass[10pt,twocolumn,letterpaper]{article}

\usepackage{cvpr}
\usepackage{times}
\usepackage{epsfig}
\usepackage{graphicx}
\usepackage{amsmath}
\usepackage{amssymb}

\usepackage{mystyle}
\usepackage{booktabs}

\usepackage[breaklinks=true,letterpaper=true,colorlinks,bookmarks=false]{hyperref}

\cvprfinalcopy 

\def\cvprPaperID{4029}

\ifcvprfinal\fi
\begin{document}

\graphicspath{{images/}}

\title{T-Net: Parametrizing Fully Convolutional Nets with a Single High-Order Tensor}

\author{Jean Kossaifi\thanks{Equal contribution.} \qquad Adrian Bulat\footnotemark[1]  \qquad Georgios Tzimiropoulos \qquad Maja Pantic
        \vspace{5pt}\\
		Samsung AI Center, Cambridge\\
		United Kingdom\\
		{\tt\small \{j.kossaifi, adrian.bulat, georgios.t, maja.pantic\}@samsung.com}
		}

\maketitle
\thispagestyle{empty}

\begin{abstract}
Recent findings indicate that over-parametrization, while crucial for successfully training deep neural networks, also introduces large amounts of redundancy. Tensor methods have the potential to efficiently parametrize over-complete representations by leveraging this redundancy. In this paper, we propose to fully parametrize Convolutional Neural Networks (CNNs) with a single high-order, low-rank tensor. Previous works on network tensorization have focused on parametrizing individual layers (convolutional or fully connected) only, and perform the tensorization layer-by-layer separately. In contrast, we propose to jointly capture the full structure of a neural network by parametrizing it with a \emph{single} high-order tensor, the modes of which represent each of the architectural design parameters of the network (e.g. number of convolutional blocks, depth, number of stacks, input features, etc). This parametrization allows to regularize the whole network and drastically reduce the number of parameters. Our model is end-to-end trainable and the low-rank structure imposed on the weight tensor acts as an implicit regularization. We study the case of networks with rich structure, namely Fully Convolutional Networks (FCNs), which we propose to parametrize with a single $8$\myth--order tensor. We show that our approach can achieve superior performance with small compression rates, and attain high compression rates with negligible drop in accuracy for the challenging task of human pose estimation.
\end{abstract}

\section{Introduction}

For a wide range of challenging tasks, including recognition \cite{krizhevsky2012imagenet, simonyan2014very,he2016deep}, detection \cite{ren2015faster}, semantic segmentation \cite{long2015fully, he2017mask} and human pose estimation \cite{newell2016stacked}, the state-of-the-art is currently attained with Convolutional Neural Networks (CNNs). There is evidence that a key feature behind the success of these methods is over-parametrization, which helps find good local minima \cite{du2018power, soltanolkotabi2018theoretical}. However, at the same time, over-parametrization leads to a great amount of redundancy, and from a statistical perspective, it might hinder generalization because it excessively increases the number of parameters. Furthermore, models with an excessive number of parameters have increased storage and computation requirements, rendering them problematic for deployment on devices with limited computational resources. This paper focuses on a novel way of leveraging the redundancy in the parameters of CNNs by jointly parametrizing the whole network using tensor methods. 

There is a significant amount of recent work on using tensors to reduce the redundancy and improve the efficiency of CNNs, mostly focusing on re-parametrizing individual layers. For example, \cite{tai2015convolutional, yong2015compression} treat a convolutional layer as a 4D tensor and then compute a decomposition of the 4D tensor into a sum of a small number of low-rank tensors. Similarly, \cite{novikov2015tensorizing} proposes tensorizing the fully-connected layers. The bulk of these methods focus on tensorizing individual layers only, and perform the tensorization layer-by-layer disjointly, usually by applying a tensor decomposition to the pre-trained weights and then fine-tuning to compensate for the performance loss. For example, \cite{tai2015convolutional} tensorizes the second convolutional layer of AlexNet \cite{krizhevsky2012imagenet}. 

Our paper primarily departs from prior work by using a single high-order tensor to parametrize the whole CNN as opposed to using different tensors to parametrize the individual layers. In particular, we propose to parametrize the network with a single high-order tensor, each dimension of which represents a different architectural design parameter of the network. For the case of Fully Convolutional Networks (FCNs) with an encoder-decoder structure considered herein (see also Fig.~\ref{fig:overall-architecture}), each dimension of the $8-$dimensional tensor represents a different architectural design parameter of the network such as the number of (stacked) FCNs used, the depth of each network, the number of input and output features for each convolutional block and the spatial dimensions of each of the convolutional kernels. By modelling the whole FCN with a single tensor, our approach allows for learning correlations between the different tensor dimensions and hence to fully capture the structure of the network. Moreover, this parametrization implicitly regularizes the whole network and drastically reduces the number of parameters by imposing a low-rank structure on that tensor. Owing to these properties, our framework is much more general and flexible compared to prior work offering increased accuracy and high compression rates.
In summary, the \textbf{contributions} of this work are:
\begin{itemize}
    \item 
    We propose using a single high-order tensor for whole network tensorization and applying it for capturing the rich structure of Fully Convolutional Networks. Our end-to-end trainable approach allows for a wide spectrum of network decompositions and compression rates which can be chosen and optimized for a particular application.
    \item
    We show that for a large range of compression rates (both high and low), our method preserves high accuracy. Compared to prior work based on tensorizing individual convolutional layers, our method consistently achieves higher accuracy, especially for the case of high compression rates. In addition, we show that, for lower compression rates, our method outperforms the original uncompressed network.
    \item
    We illustrate the favorable properties of our method by performing a large number of experiments and ablation studies for the challenging task of human pose estimation. The experiments shed light on several interesting aspects of our method including the effect of varying the rank for each mode of the tensor, as well as the decomposition method used. We further validate our conclusions on a different dense prediction task, namely semantic facial part segmentation.
\end{itemize}

\section{Related Work}

In this section, we review related work, both for tensor methods and human pose estimation. 

\paragraph{Tensor methods} offer a natural extension of traditional algebraic methods to higher orders. For instance, Tucker decomposition can be seen as a generalization of PCA to higher dimensions \cite{tensor_decomposition_kolda}. Tensor decompositions have wide-reaching applications, including learning a wide range of probabilistic latent-variable models~\cite{decomposition_latent_anandkumar}.
Tensor methods have been recently applied to deep learning, for instance, to provide a theoretical analysis of deep neural nets \cite{expressive_deep_tensor}. New layers were also proposed, leveraging tensor methods. \cite{kossaifi2017tensor} proposes tensor contraction layers to reduce the dimensionality of activation tensors while preserving their multi-linear structure. Tensor regression layers  \cite{kossaifi2018tensor} express outputs through a low-rank multi-linear mapping from a high-order activation tensor to an output tensor of arbitrary order.

A lot of existing work has been dedicated to leveraging tensor decompositions in order to re-parametrizing existing layers, either to speed up computation or to reduce the number of parameters. Separable convolutions, for instance, can be obtained from existing ones by applying CP decomposition to their kernel. The authors in \cite{lebedev2014speeding} propose such parametrization of individual convolutional layers using CP decomposition with the goal of speeding them up. Specifically, each of the 4D tensors parametrizing the convolutional layers of a pre-trained network are decomposed into a sum of rank--\(1\) tensors using CP decomposition. The resulting factors are used to replace each existing convolution with a series of \(4\) convolutional layers with smaller kernels. The network is then fine-tuned to restore performance. 
\cite{yong2015compression} proposes a similar approach but uses Tucker decomposition instead of CP to decompose the convolutional layers of a pre-trained network, before fine-tuning to restore the performance. Specifically, Tucker decomposition is applied to each convolutional kernel of a pre-trained network, on two of the modes (input and output channel modes). The resulting network is fine-tuned to compensate for the drop in performance induced by the compression.

In \cite{astrid2017cp}, the layers of deep convolutional neural networks are also re-parametrized using CP decomposition, optimized using the tensor power method. The network is then iteratively fine-tuned to restore performance.
Similarly, \cite{tai2015convolutional} proposes to use tensor decomposition to remove redundancy in convolutional layers and express these as the composition of two convolutional layers with less parameters. Each 2D filter is approximated by a sum of rank--\(1\) matrices. Thanks to this restricted setting, a closed-form solution can be readily obtained with SVD. This is done for each convolutional layer with a kernel of size larger than 1.
While all these focus of convolutional layers, other types of layers can also be parametrized.
For instance, \cite{novikov2015tensorizing} uses the Tensor-Train (TT) format \cite{oseledets2011tensor} to impose a low-rank tensor structure on the weights of the fully-connected layers. Tensorization of generative adversarial networks \cite{tensor-gan} and sequence models~\cite{tensor-rnn} have  been also proposed.

The work of \cite{chen2017sharing} proposes a new residual block, the so-called \emph{collective residual unit} (CRU), which is obtained by applying a generalized block term decomposition to the last two modes of a 4\myth--order tensor obtained by stacking the convolutional kernels of several residual units. Similarly to existing works, each of the CRUs is parametrized individually. \cite{yang2016deep} leverages tensor decomposition for multi-task learning to allow for weight sharing between the fully-connected and convolutional layers of two or more deep neural networks.

Overall, to the best of our knowledge, our work is the first to propose an end-to-end trainable architecture, fully parametrized by a \emph{single} high order low-rank tensor.

\paragraph{Other methods for network decomposition.} 
There are also other methods, besides tensor-based ones, for reducing the redundancy and number of parameters in neural networks. A popular approach is quantization which is concerned with quantizing the weights and/or the features of a network \cite{tung2018clip,wu2016quantized,zhou2018explicit,tang2018quantized, courbariaux2015binaryconnect,rastegari2016xnor}. Quantization methods should be considered orthogonal to tensor methods as one could apply them to the output of tensor decompositions, too. Similarly, complementary to our work should be considered methods for improving the efficiency of neural networks using weight pruning \cite{li2017pruning, he2017channel}.

More related to our work are hand-crafted decomposition methods such as MobileNet~\cite{howard2017mobilenets} and Xception~\cite{chollet2017xception} which decompose $3 \times 3$ convolutions using efficient depth-wise and point-wise convolutions. We compare our methods with MobileNet, the method of choice for improving the efficiency of CNNs, and show that our approach outperforms it by large margin. 

\paragraph{Human pose estimation.} CNN--based methods have recently produced results of remarkable accuracy for the task of human pose estimation, outperforming traditional methods by large margin \cite{toshev2014deeppose,tompson2014joint,pfister2015flowing, bulat2016human,newell2016stacked, wei2016convolutional}. Arguably, one of the most widely used architectures for this task is the stacked HourGlass (HG) network proposed by \cite{newell2016stacked}. An HG is an encoder-decoder network with skip connections between the encoder and the decoder, suitable for making predictions at a pixel level in a fully convolutional manner. \cite{newell2016stacked}~uses a stack of 8 of these networks to achieve state-of-the-art performance on the MPII dataset \cite{andriluka20142d}. The architecture is shown in Fig.~\ref{fig:overall-architecture}. In this work, we choose tensorizing the HG network primarily because of its rich structure which makes it suitable to model it with a high-order tensor. We note that the aim of this work is not to produce state-of-the-art results for the task of human pose estimation but to show the benefits of modelling a state-of-the-art architecture with a single high-order tensor.

 \section{Mathematical background}\label{sec:math-background}

In this section we first introduce some mathematical background regarding the notation and tensor methods used in this paper.

\paragraph{Notation.} We denote vectors (1\myst--order tensors) as \(\myvector{v}\), matrices (2\mynd--order tensors) as \(\mymatrix{M}\), and tensors of order 3 or greater as  \(\mytensor{X}\) . We denote element \((i_0, i_1, \cdots, i_N)\) of a tensor as \(\mytensor{X}_{i_0, i_1, \cdots, i_N} \,\text{ or } \, \mytensor{X}(i_0, i_1, \cdots, i_N)\). A colon is used to denote all elements of a mode, e.g. the mode--1 fibers of \(\mytensor{X}\) are denoted as \(\mytensor{X}(:, i_2, i_3, \cdots, i_N)\). 
Finally, for any \(i, j \in \myN, \myrange{i}{j}\) denotes the set of integers \(\{ i, i+1, \cdots , j-1, j\}\).

\paragraph{Mode--\(n\) unfolding} of a tensor
	\( \mytensor{X} \in \myR^{I_0 \times I_1 \times \cdots \times I_N}\),
	is a matrix \(\mymatrix{X}_{[n]} \in \myR^{I_n, I_M}\), 
	with \(M = \prod_{\substack{k=0,\\k \neq n}}^N I_k\),
	defined by the mapping from element
	\( (i_0, i_1, \cdots, i_N)\) to \((i_n, j)\), with 
	\(
	j = \sum_{\substack{k=0,\\k \neq n}}^N i_k \times \prod_{\substack{m=k+1,\\ m \neq n}}^N I_m 
	\).

\paragraph{Mode-n product.}
For a tensor \(\mytensor{X} \in \myR^{I_0 \times I_1 \times \cdots \times I_N}\) and a matrix \( \mymatrix{M} \in \myR^{R \times I_n} \), the n-mode product of a tensor 
	is a tensor of size 
	\(\left(I_0 \times \cdots \times I_{n-1} \times R \times I_{n+1} \times \cdots \times I_N\right)\) 
	and can be expressed using the unfolding of \(\mytensor{X}\) 
	and the classical dot product as \(
		\mytensor{X} \times_n \mymatrix{M} = \mymatrix{M} \mytensor{X}_{[n]} \in \myR^{I_0 \times \cdots \times I_{n-1} \times R \times I_{n+1} \times \cdots \times I_N}\)

\paragraph{Tensor diagrams.} While \(2\mynd\)--order tensors can easily be depicted as rectangles and \(3\myrd\)--order tensors as cubes, it is impractical to represent high-order tensors in such way. We instead use tensor diagrams, which are undirected graphs where the vertices represent tensors. The \emph{degree} of each vertex (i.e. the number of edges originating from this circle) specifies the order of the corresponding tensor. Tensor contraction over two modes is then represented by simply linking together the two edges corresponding to these two modes. Fig.~\ref{fig:tucker-form} depicts the Tucker decomposition (i.e. contraction of a core tensor with factor matrices along each mode) of an 8\myth--order tensor with tensor diagrams.
 \section{T-Net: Fully-tensorized FCN architecture}\label{sec:method}

In this section, we introduce our fully-tensorized method by first introducing the architecture before detailing the structure of the parametrization weights.
\begin{figure*}
    \centering
    \includegraphics[width=1\linewidth]{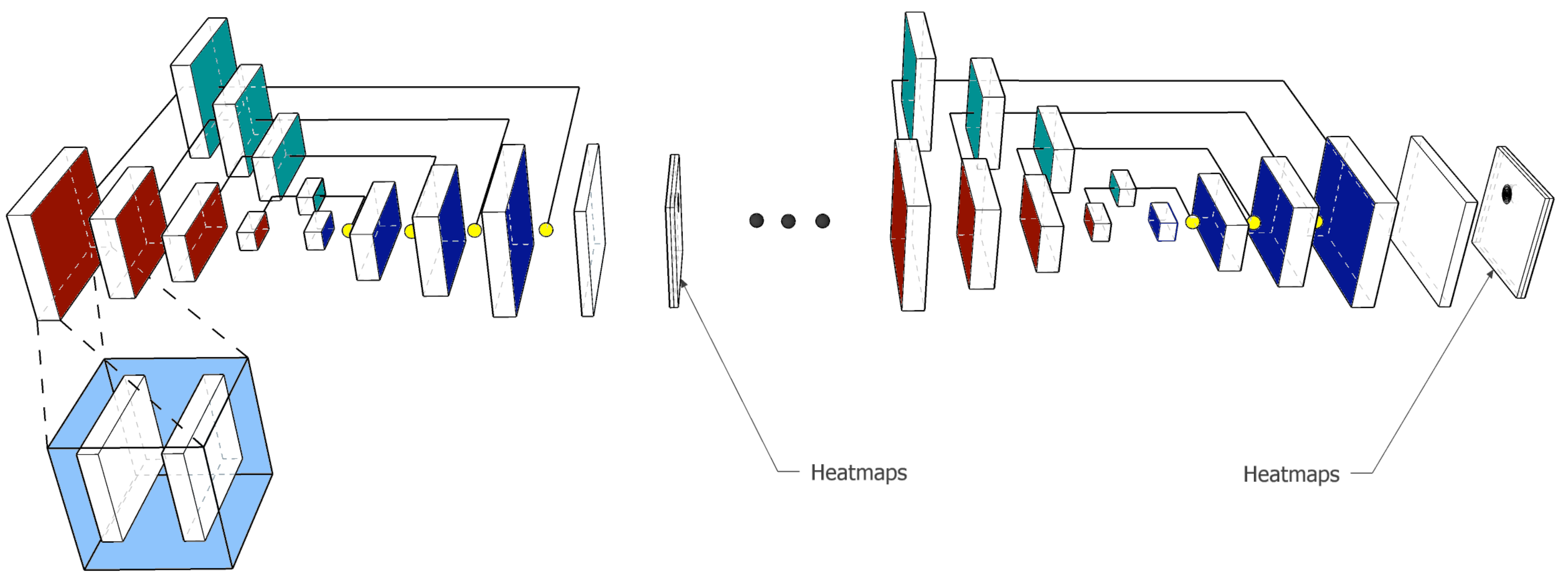}
    \caption{\textbf{Overall network architecture.} Each block in the fully convolutional network is a basic-block module~\cite{he2016deep} (\emph{blue insert}), containing \emph{b\mysub{depth}} (by default \(2\)) convolutional layers with \(3\times3\) kernels followed by BatchNorm and ReLU. For all experiments, unless explicitly stated otherwise, we used a stack of \(4\) sub-networks with \(3\) pathways each: downsampling/encoder (red blocks), upsampling/decoder (dark blue) and skip connection (cyan). Yellow dots are element-wise sums.}
    \label{fig:overall-architecture}
\end{figure*}

\subsection{FCN tensorization}

In this section, we describe how to tensorize the stacked HourGlass (HG) architecture of \cite{newell2016stacked}. The HG  has a number of design parameters namely the number of (stacked) HGs, the depth of each HG, the three signal pathways of each HG (skip, downsample and upsample), the number of convolutional layers in each residual block (\myie the depth of each block), the number of input and output features of each block and finally, the spatial dimensions of each of the convolutional kernels. 

\begin{figure}
    \centering
    \includegraphics[width=1\linewidth]{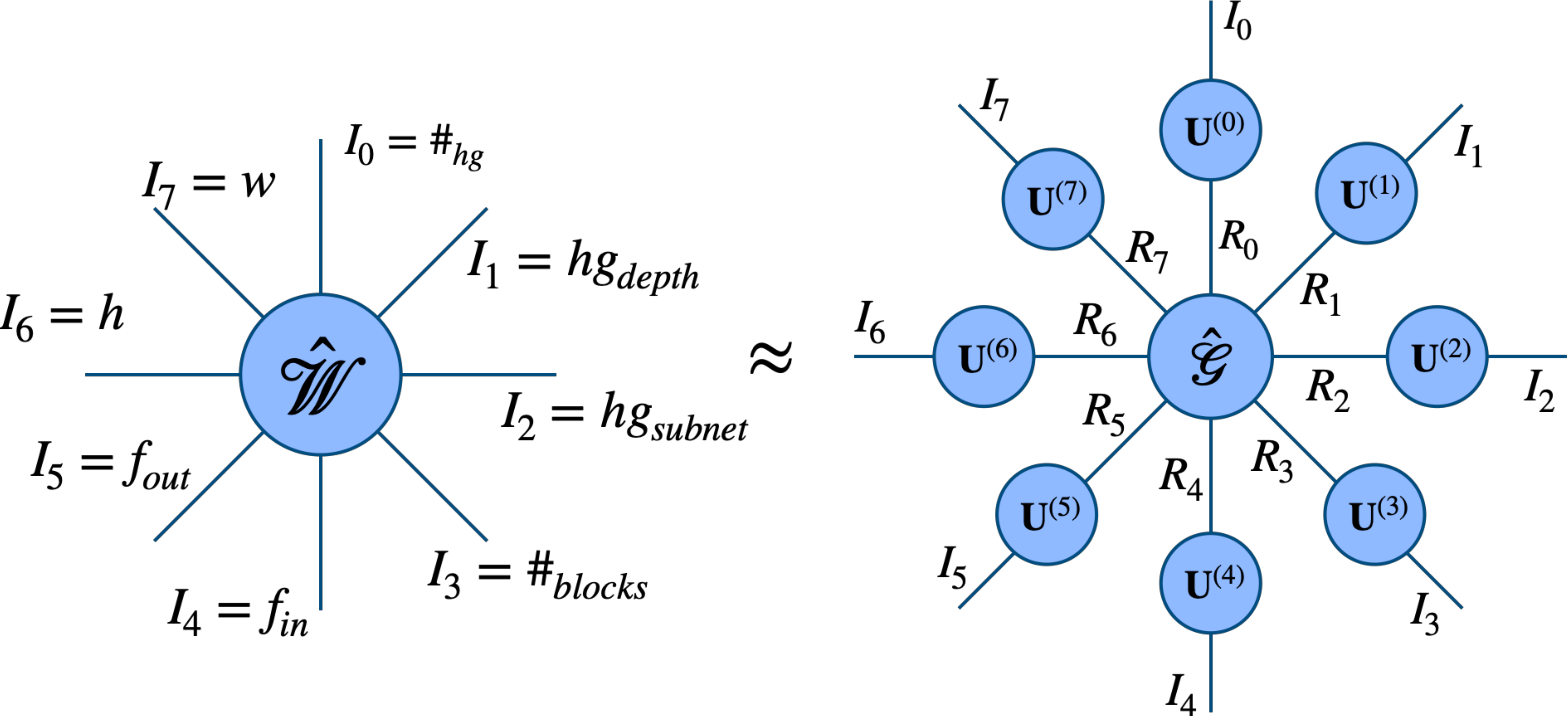}
    \caption{\textbf{Tensor diagram} of the Tucker form of the weight tensor \(\mytensor{W}\) parametrizing our model.}
    \label{fig:tucker-form}
    \vspace{-10pt}
\end{figure}

To facilitate the tensorization of the whole network, we used a modified HG architecture in which we replaced all the residual modules with the basic block introduced by \cite{he2016deep}, maintaining the same number of input and output channels throughout the network. We made the encoder and the decoder symmetric, with 4 residual modules each. Figure~\ref{fig:overall-architecture} illustrates the modified HG architecture. We note that from an accuracy perspective, this modification performs (almost) the same as the original HG proposed in \cite{newell2016stacked}.

From the network described above, we derive the high-order tensor for the proposed Tensorized-Network (T-Net) as follows: all weights of the network are parametrized by a \emph{single} \(8\myth\)--order tensor \(\mytensor{W}  \in \myR^{I_0 \times I_1 \times \cdots \times I_7}\), the modes of which correspond to the number of HGs (\(I_0 = \) \emph{\#hg}), the depth of each HG (\(I_1 = \) \emph{hg\mysub{depth}}), the three signal pathways (\(I_2 = \) \emph{hg\mysub{subnet}}), the number of convolutional layers per block (\(I_3 = \) \emph{b\mysub{depth}}), the number of input features (\(I_4 = \) \emph{f\mysub{in}}), the number of output features (\(I_5 = \) \emph{f\mysub{out}}), and finally the height (\(I_6 = \) \emph{h}) and width (\(I_7 = \) \emph{w}) of each of the convolutional kernels. 

\subsection{T-Net variants}
Based on the previous parametrization of the network, we can add various low-rank constraints on the weight tensor, leading to variants of our method.

\paragraph{Tucker T-Net.} The Tucker form of our model expresses the constructed 8\myth--order tensor \(\mytensor{W} \) as a rank--\((R_0, \cdots, R_7)\) Tucker tensor, composed of a low rank core 
\(\mytensor{G} \in \myR^{R_0 \times R_1 \times \cdots \times R_7}\) 
along with projection factors
\( \left( \mymatrix{U}^{(0)}, \cdots,\mymatrix{U}^{(7)} \right) \), with \(\mymatrix{U}^{(k)} \in \myR^{R_k, I_k}, k \in \myrange{0}{7}\).  This allows us to write the network's weight tensor in a decomposed form as:
\begin{align}
\mytensor{W} &=
\mytensor{G} \times_0 \mymatrix{U}^{(0)} 
		  \times_1  \mymatrix{U}^{(2)} \times
		  \cdots
          \times_7 \mymatrix{U}^{(7)}\\
          \nonumber 
          &= \mytucker{\mytensor{G}}{\mymatrix{U}^{(0)}, \cdots, \mymatrix{U}^{(7)}} 
\end{align}
See also Fig. \ref{fig:tucker-form} for a tensor diagram of the Tucker form of the weight tensor. Note that the CP decomposition is the special case of the Tucker decomposition, where the core is super-diagonal.

\paragraph{MPS T-Net.} The Matrix-Product-State (MPS) form (also known as \emph{tensor-train}~\cite{oseledets2011tensor}) of our model expresses the constructed 8\myth--order weight tensor \(\mytensor{W} \) as a series of third-order tensors (the \emph{cores}) and allows for especially large space-savings. In our case, given \(\mytensor{W} \in \myR^{I_0 \times I_1 \times \cdots \times I_7} \), 
we can decompose it into a rank \((R_0, R_1, \cdots R_8)\)--MPS as a series of third-order cores \(\mytensor{G}_0 \in \myR^{R_0, I_0, R_1}, \mytensor{G}_1 \in \myR^{R_1, I_1, R_2}, \cdots, \mytensor{G}_7 \in \myR^{R_7, I_7, R_8} \). The boundary conditions dictate \(R_0 = R_8 = 1\). In terms of individual elements, we can then write, for any \(i_0 \in \myrange{0}{I_0}, i_1 \in \myrange{0}{I_1}, \cdots, i_7  \in \myrange{0}{I_7}\):
\begin{equation}
    \mytensor{W}(i_0, i_1, \cdots, i_7) =
        \underbrace{\mytensor{G}_0[i_0]}_{1 \times R_1}
        \times
        \underbrace{\mytensor{G}_1[i_1]}_{R_1 \times R_2}
        \times 
        \cdots
        \times
        \underbrace{\mytensor{G}_7[i_7]}_{R_{7} \times 1} \nonumber
\end{equation}

\begin{figure}
    \centering
    \includegraphics[width=1\linewidth]{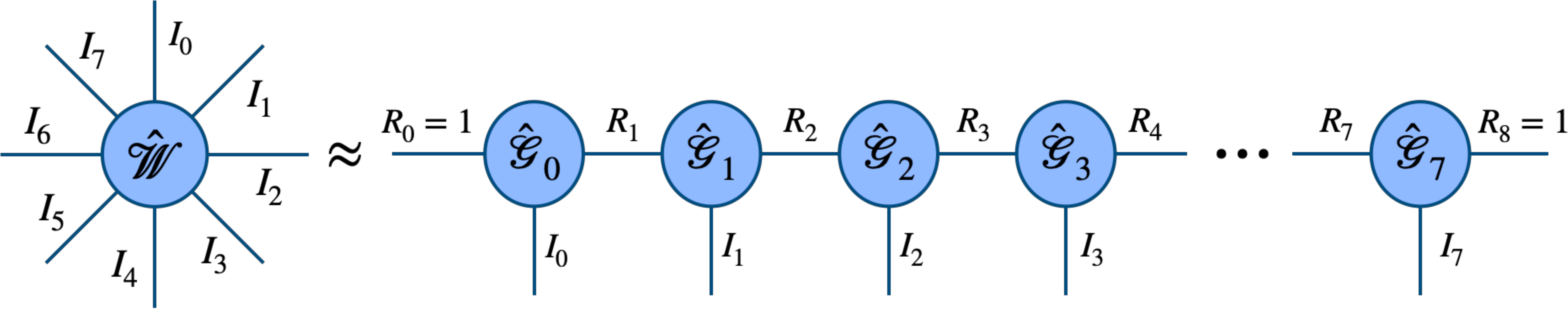}
    \caption{\textbf{Tensor diagram of the MPS/TTrain form of the weight tensor \(\mytensor{W}\)}. Note the \emph{train}--like shape from which the method takes its name, as well as the boundary conditions (\(R_0 = R_8 = 1\)).}
    \label{fig:mps-form}
\end{figure}

\subsection{Parameter analysis}

This section compares the number of parameters of our model which parameterizes the whole weight tensor with a single high-order tensor with methods based on layer-wise decomposition (e.g. \cite{yong2015compression, lebedev2014speeding}). Considering a Tucker rank--\(R_0, R_1, \cdots, R_7\) of the weight tensor parametrizing the whole network, the resulting number of parameters is:

\begin{equation}
N_{T-Net} = \prod_{k=0}^7 R_k + \sum_{k=0}^7 R_k \times I_k.
\end{equation}

Compressing each of the \(N_{\text{conv}}\) convolutional layer separately \cite{yong2015compression}, with a rank \(R_{4}\) and \(R_{5}\) for the number of input and output features, respectively,  and writing \(N_{\text{conv}} = \prod_{k=0}^4 I_k\), we obtain the total number of parameters: 
\begin{equation}
N_{\text{conv}} \times \left( R_{4} \times R_{5} \times I_6 \times I_7  + R_{4}\times I_4 + R_{5} I_5 \right).
\end{equation}

In comparison, our model, with the same ranks \(R_4\) and \(R_5\) imposed on the number of features, would only have \( N_{\text{conv}} \times \left( R_{4} \times R_{5} \times 3 \times 3 \right) + R_{4}\times I_4 + R_{5} I_5 \) parameters. In other words, our model has \(\left(\prod_{k=0}^4 I_k - 1\right) \left( R_{4}\times I_4 + R_{5} I_5\right)\) parameters less than a corresponding layer-wise decomposition.

\paragraph{Speeding up the convolutions.} When parametrized using a CP or Tucker decomposition, a convolutional layer can be efficiently replaced by a series of convolutions with smaller kernels \cite{lebedev2014speeding,yong2015compression}, thus allowing for large computational speedups. This efficient re-parametrization also applies to our model. To see this, given the weight tensor \(\mytensor{W} \in \myR^{I_0 \times I_1 \times \cdots \times I_7} \) of our Tucker T-Net, we have
\[
\mytensor{W} = \mytensor{G} \times_0 \mymatrix{U}^{(0)} \times_1  \mymatrix{U}^{(1)} \times \cdots \times_7 \mymatrix{U}^{(7)}.
\]

For any \(i_0, i_1, i_2, i_3 \in (I_0, I_1, I_2, I_3)\), let us denote  
\(
\mytensor{\tilde K} = \mytensor{W}(i_0, i_1, i_2, i_3, \mycolon, \mycolon, \mycolon, \mycolon)
\), corresponding to one of the convolutional kernels of the T-Net. By re-arranging the terms, and considering the partially contracted core, we can write: 
\[
\mytensor{\tilde K}(s, t, j, k) = 
\sum_{r_4=0}^{R_4} \sum_{r_5=0}^{R_5} \mytensor{C}(r_4, r_5, j, k)  \mymatrix{U}^{(4)}(s, r_4) \mymatrix{U}^{(5)}(t, r_5)
\]
with \(\mytensor{C} = \mytensor{C}_{i_0, i_1, i_2, i_3, \mycolon, \mycolon, \mycolon, \mycolon} \in \myR^{(I_4, I_5, I_6, I_7)}\) and
\[
\mytensor{C} = \left(\mytensor{G} \times_0 \mymatrix{U}^{(0)} \times  \cdots \times_3  \mymatrix{U}^{(3)} \times_6 \mymatrix{U}^{(6)} \times_7 \mymatrix{U}^{(7)}\right).
\]
This gives us an effective way of approximating each convolution by three smaller convolutions \cite{yong2015compression}. While getting the full speedup would require the writing of specialized CUDA kernels, some timings results with a naive implementation using PyTorch are shown in Table~\ref{tab:timing}, for a single convolutional layer with a kernel tensor of size \(128 \times 128 \times 3 \times 3\) compressed using Tucker decomposition.

\begin{table}[t]
    \begin{center}
    \resizebox{1\linewidth}{!}{
    \begin{tabular}{c|c|c|c}
         \textbf{Baseline} & \textbf{Tucker $1.37\times$} & \textbf{Tucker $2.77\times$} & \textbf{Tucker $4.17\times$}\\
         \hline
         3.79 ms. & 4.36 ms. & 2.72 ms. & \textbf{2.45 ms.}
    \end{tabular}
    }
    \end{center}
    \vspace{-8pt}
    \caption{\textbf{Timing of baseline conv. vs. naive Tucker.} Speed-up for a \(3 \times 3\) convolution preserving the number of channels and input tensor of size ($128\times64\times64$), with a batch-size $64$ . We vary the Tucker-rank and report times.}
    \label{tab:timing}
\vspace{-10pt}
\end{table} \section{Experimental Setup}
\begin{table*}[ht]
\begin{center}
\begin{tabular}{ cccc cccc | cc }
\toprule
\multicolumn{8}{c|}{\textbf{Tucker-rank}} & \textbf{Accuracy} & \textbf{Compression} \\
\cline{1-8}
\textbf{\#hg} & \textbf{hg\mysub{depth}} & \textbf{hg\mysub{subnet}} & \textbf{b\mysub{depth}}
& \textbf{f\mysub{in}} & \textbf{f\mysub{out}} & \textbf{h} & \textbf{w} & \textbf{(PCKh)} & \textbf{ratio}\\ 
\toprule
\multicolumn{8}{c|}{\textit{Original}} & 86.99\% & 1.0x \\  
\hline
\textbf{3} & 4 & 3 & 2 & 128 & 128 & 3 & 3 & 87.42\% & 1.28x \\
\textbf{2} & 4 & 3 & 2 & 128 & 128 & 3 & 3 & 86.95\% & 1.82x \\
\textbf{1} & 4 & 3 & 2 & 128 & 128 & 3 & 3 & 86.05\% & 3.03x \\
\hline
4 & \textbf{3} & 3 & 2 & 128 & 128 & 3 & 3 & 87.71\% & 1.28x \\
4 & \textbf{2} & 3 & 2 & 128 & 128 & 3 & 3 & 87.59\% & 1.82x \\
4 & \textbf{1} & 3 & 2 & 128 & 128 & 3 & 3 & 86.89\% & 3.03x \\
\hline
4 & 4 & \textbf{2} & 2 & 128 & 128 & 3 & 3 & 87.53\% & 1.43x \\
4 & 4 & \textbf{1} & 2 & 128 & 128 & 3 & 3 & 86.19\% & 2.50x \\
\hline
4 & 4 & 3 & \textbf{1} & 128 & 128 & 3 & 3 & 82.59\% & 1.82x \\
\hline
4 & 4 & 3 & 2 & \textbf{96} & \textbf{96} & 3 & 3 & 87.43\% & 1.64x \\
4 & 4 & 3 & 2 & \textbf{64} & \textbf{64} & 3 & 3 & 86.13\% & 3.03x \\
4 & 4 & 3 & 2 & \textbf{32} & \textbf{32} & 3 & 3 & 83.10\% & 6.25x \\
\hline
4 & 4 & 3 & 2 & 128 & 128 & \textbf{2} & \textbf{2} & 87.30\% & 1.98x \\
\bottomrule
\end{tabular}
\caption{\textbf{Human pose estimation task.} Study of the redundancy of each of the modes of the $8$\myth--order weight tensor. We compress one dimension at a time by reducing its corresponding rank in the Tucker tensor. Reported accuracy is in terms of PCKh.}
\label{table:redundancy-tucker}
\end{center}
\end{table*} \begin{figure*}[ht]
    \centering
    \includegraphics[width=\linewidth, trim={1cm 1cm 1cm 1cm}, clip]{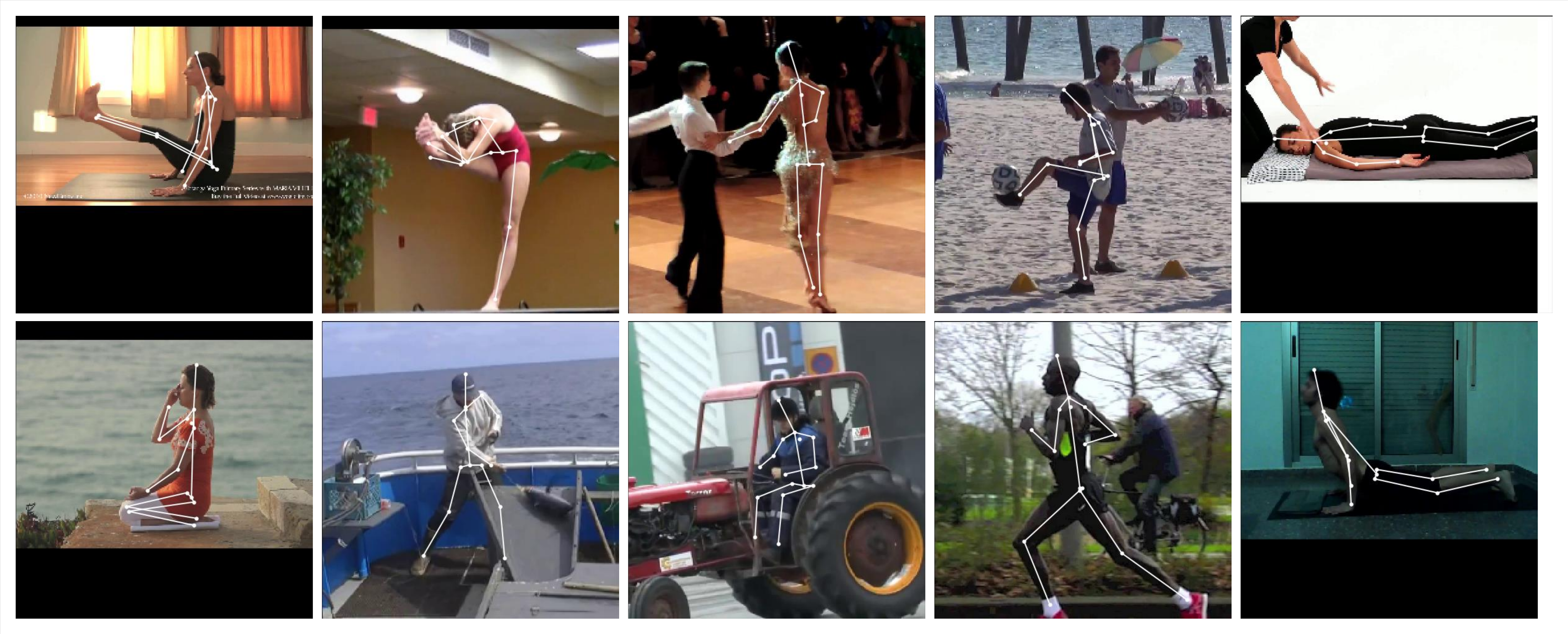}
    \caption{\textbf{Qualitative results produced by our method on MPII.}}
    \label{fig:human-pose-example}
\end{figure*}

The bulk of our experiments were conducted for the task of human pose estimation. We also validated some of our conclusions by conducting experiments for a different dense prediction task, namely facial part segmentation. 

\paragraph{Human pose estimation.} Following \cite{tompson2014joint}, we conducted experiments using the standard train and validation splits of one of the most challenging single pose human pose estimation datasets, namely MPII~\cite{andriluka20142d}. The dataset contains 22,000 images for training and another 3,000 for validation.

\paragraph{Semantic facial part segmentation.} We constructed the facial part segmentation dataset as in~\cite{bulat2017binarized}: for training, we used the 300W training dataset (more than 3,000 images) and for testing  the whole 300W competition test set (600 images) \cite{sagonas2013300}.

\paragraph{Implementation details}
We used a stacked HG architecture with the following architectural parameters: \emph{\#hg} = 4, \emph{hg\mysub{depth}} = 4,  \emph{hg\mysub{subnet}}=3, \emph{b\mysub{depth}} = 2, \emph{f\mysub{in}}=128, \emph{f\mysub{out}}=128, and \emph{h} = \emph{w} = 3.  This resulted in a \(8^{\text{th}}\)--order tensor of size \(4 \times 4 \times 3 \times 2 \times 128 \times 128 \times 3 \times 3\). 

For the uncompressed baseline network, we reduced the number of its parameters by simply decreasing the number of channels in each residual block, varying it from 128 to 64. By doing so, (as opposed to reducing the number of stacks), we maintain all the architectural advantages offered by the stacked HG architecture and ensure a fair comparison with the proposed tensorized network.

\paragraph{Training.} All models were trained for 110 epochs using \mbox{RMSprop}~\cite{tieleman2012lecture}. The learning rate was varied from $2.5e-4$ to $1e-6$ using a Multi-Step fixed scheduler. During training, we randomly augmented the data using: rotation ($-25^\circ$ to $25^\circ$ for human pose and $-40^\circ$ to $40^\circ$ for face part segmentation), scale distortion ($0.75$ to $1.25$), horizontal flipping and color jittering. 

All experiments were run on a single NVIDIA TITAN V GPU. All networks were implemented using \emph{PyTorch}~\cite{pytorch}. \emph{TensorLy}~\cite{tensorly} was used for all tensor operations.

\begin{table*}[ht]
\begin{center}
\begin{tabular}{ ccccc }
\toprule
 \textbf{Method} &  \textbf{Parameters} & \textbf{Compression ratio} & \textbf{Accuracy} \\ 
 \midrule
 Uncompressed Baseline & full, f\mysub{in}=f\mysub{out}=128 & 1x & 87\% \\
 \midrule 
 Trimmed Baseline & f\mysub{in}=f\mysub{out}=112 & 1.3x & 86.9\% \\  
 Trimmed Baseline & f\mysub{in}=f\mysub{out}=92 & 2x & 85.9\% \\
 Trimmed Baseline & f\mysub{in}=f\mysub{out}=64 & 4x & 84.5\% \\ 
 Trimmed Baseline & hg\_depth=3 & 1.3x & 86.79\% \\  
 Trimmed Baseline & hg\_depth=2 & 1.8x &  86.82\% \\
 Trimmed Baseline & hg\_depth=1 & 3.0x &  85.30\% \\ 
 \midrule 
 MobileNet-\cite{mobilenets} & f\mysub{in}=f\mysub{out}=194 & 3.6x & 84.3\% \\
 MobileNet-\cite{mobilenets} & f\mysub{in}=f\mysub{out}=160 & 5.4x & 82.7\% \\
 \midrule
  \cite{yong2015compression} & rank--(128, 128, 2, 2) & 1.4x & 84.9\% \\
  \cite{yong2015compression} & rank--(96, 96, 3, 3) & 1.3x & 86.8\% \\
  \cite{yong2015compression} & rank--(64, 64, 3, 3) & 2.3x & 86.4\% \\
  \cite{yong2015compression} & rank--(32, 32, 3, 3) & 4.7x & 85.3\% \\
  \cite{yong2015compression} & rank--(16, 16, 3, 3) & 6.9x & 83.7\% \\
  \midrule
  \textbf{Tucker T-Net [Ours]} & rank--\((4, 3, 3, 2, 110, 110, 3, 3)\) & 1.7x & \textbf{87.5\%} \\
  \textbf{Tucker T-Net [Ours]} & rank--\((4, 4, 2, 2, 110, 110, 3, 3)\) & 1.8x & \textbf{87.4\%} \\
  \textbf{Tucker T-Net [Ours]} & rank--\((3, 3, 3, 2, 110, 110, 2, 2)\) & 3.7x & \textbf{87.1\%} \\
 \textbf{Tucker T-Net [Ours]} & rank--\((3, 2, 3, 2, 96, 96, 3, 3)\) & 3.4x & \textbf{86.7\%} \\
  \textbf{Tucker T-Net [Ours]} & rank--\((3, 3, 2, 2, 80, 80, 3, 3)\) & 4.2x & \textbf{86.3\%} \\
  \textbf{Tucker T-Net [Ours]} & rank--\((2, 2, 2, 2, 96, 96, 3, 3)\) & 5.2x & \textbf{86.0\%} \\
 \midrule
  \textbf{MPS T-Net [Ours]} & rank--\((1, 4, 4, 12, 24, 110, 9, 3, 1)\) & 7.4x & \textbf{85.5\%} \\
 \bottomrule
\end{tabular}
\caption{\textbf{Human pose estimation task}. Comparison between T-Net and various baselines and state-of-the-art methods. Accuracy is reported in terms of PCKh. For the tensor decomposition-based methods, we report the rank, and for the others, the number of channels in the convolutional layers.}
\label{table:results}
\end{center}
\end{table*} 
\paragraph{Performance measures.} For the human pose estimation experiments, we report accuracy in terms of PCKh~\cite{andriluka20142d}. For facial part segmentation, we report segmentation accuracy using the mean accuracy and mIOU metrics \cite{long2015fully}. Finally, we measure the parameter savings using the \(\text{compression ratio} = \frac{\text{uncompressed}}{\text{compressed}}\), defined as the total number of parameters of the uncompressed network divided by the number of parameters of the compressed network.

\section{Results} \label{ssec:results}

This section offers an in-depth analysis of the performance and accuracy of the proposed T-Net. Our main results are that the proposed approach: i) outperforms the layer-wise decomposition of \cite{yong2015compression} and \cite{lebedev2014speeding}, which are the most closely related works to our method; ii) outperforms the uncompressed, original network for low compression rates; iii) achieves consistent compression ratios across arbitrary dimensions and iv) outperforms MobileNet~\cite{mobilenets} by large margin. Finally, we further validate some of these results for the task of semantic facial part segmentation.

All results reported were obtained by fine-tuning our networks in an end-to-end manner from a pre-trained uncompressed original network. We were able to reach the same level of accuracy when training from scratch, though this required training for more iterations. In contrast, we found that when trained from scratch, the layer-wise method of \cite{yong2015compression} reaches sub-par performance, as also reported in their paper.

\begin{table*}[ht]
\begin{center}
\begin{tabular}{ ccccc }
\toprule
 \textbf{Method} &  \textbf{Parameters} & \textbf{Compression ratio} & \textbf{mIOU} & \textbf{mAcc} \\ 
 \midrule
 Uncompressed baseline & full, f\mysub{in}=f\mysub{out}=128 & 1x & 76.02\%  & 97.31\% \\
 \midrule
 \textbf{T-Net [Ours]} & Tucker--\((3, 2, 3, 2, 96, 96, 3, 3)\) & 3.38x & \textbf{76.01\%}  & \textbf{97.29\%} \\
 \textbf{T-Net [Ours]} & Tucker--\((2, 2, 2, 2, 64, 64, 3, 3)\) & 6.94x & \textbf{75.57\%}  & \textbf{97.01\%} \\
 \bottomrule
\end{tabular}
\caption{\textbf{Facial part segmentation task}. Comparison between T-Net and a network with the same architecture and number of features as the compressed one. Our approach is able to retain a high accuracy even at high compression rates (up to 7x).}
\label{table:results-segmentation}
\end{center}
\end{table*}

\begin{figure*}[t]
    \centering
    \includegraphics[width=\linewidth, trim={1cm 1cm 1cm 1cm}, clip]{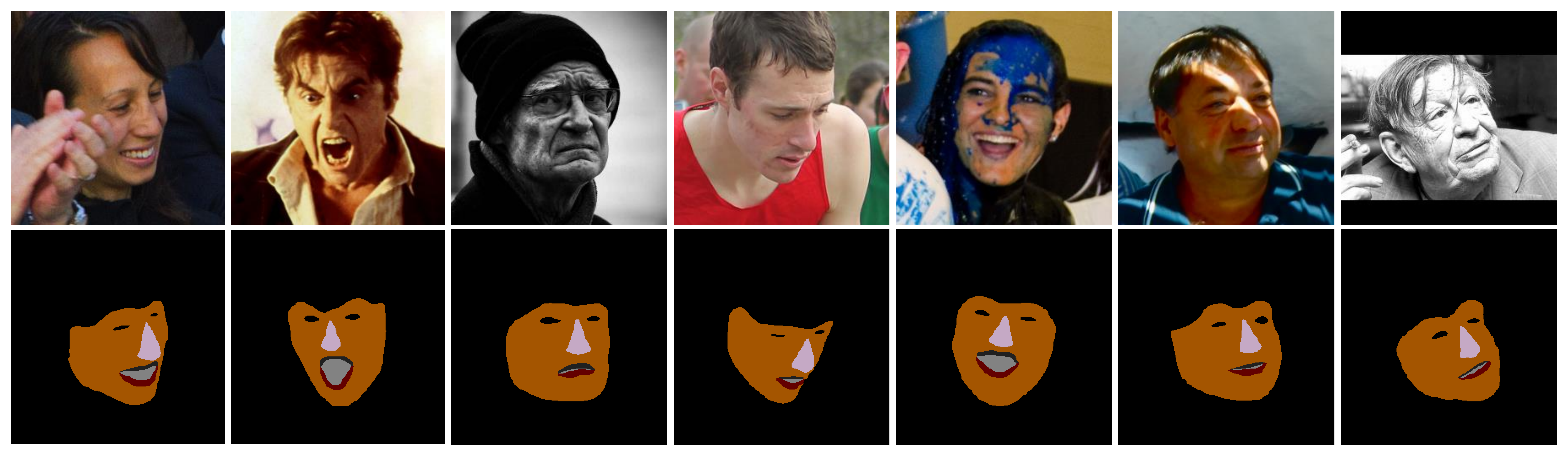}
    \caption{\textbf{Qualitative results produced by our method on the facial part segmentation task.}}
    \label{fig:face-parsing-example}
\end{figure*}

\subsection{Redundancy of the weight tensor}
In order to better understand the compressibility of each mode of the weight tensor, we first investigate the redundancy of each of the modes of the tensor by compressing \emph{only one} of the modes at a time. Table~\ref{table:redundancy-tucker} shows the accuracy (PCKh) as well as the compression ratio obtained by compressing one of the modes, corresponding respectively to the number of HGs (\emph{\#hg}), the depth of each HG (\emph{hg\mysub{depth}}), the three pathways of each HG (\emph{hg\mysub{subnet}}), the number of convolutional layers per blocks (\emph{b}\mysub{depth}) and, finally, the number of input features (\emph{f\mysub{in}}), output features (\emph{f\mysub{out}}), height (\emph{h}) and the width (\emph{w}) of each of the convolutional kernels. The results are shown along with the performance of the original uncompressed network. We observe that by taking advantage of the redundancy at network-level (as opposed to \cite{lebedev2014speeding, yong2015compression} which compress individual layers), the proposed approach is able to effectively compress across arbitrary dimensions for large compression ratios while maintaining similar, or even in some cases higher, accuracy than that of the original uncompressed network.

\subsection{Performance of the T-Net}
Based on the insights gained from the previous experiment, we selected promising configurations and compressed over multiple dimensions simultaneously. We then compared these configurations with baseline and state-of-the-art methods. The results can be seen in Table~\ref{table:results}. 

\textbf{Compression vs. trimming.} The obvious comparison is between T-Net and the original baseline network, ``compressed'' by trimming it, reducing the number of parameters to match the compression ratio achieved by T-Net.

\textbf{Comparison with efficient architectures.} A natural question is whether T-Net performs favourably when compared to architectures designed for efficiency. To answer this, we performed a comparison with MobileNet~\cite{mobilenets}, for which we adjusted the number of channels of the convolutional layers in order to vary the number of parameters and obtain comparable compression ratios.

\textbf{Comparison with the state-of-the-art.} We also compared 
with the layer-wise decomposition method of~\cite{yong2015compression}. 

We firstly observe that by just reducing the number of channels in the original network, a significant drop in performance can be noticed. Secondly, our method consistently outperforms~\cite{yong2015compression} across the whole spectrum of compression ratios. This can be seen by comparing the accuracy provided for any compression ratio for~\cite{yong2015compression} with the accuracy of the closest but higher compression ratio for our method (for example, compare $2.33$x for~\cite{yong2015compression} with $3.67$x for our method). Our method always achieves higher accuracy even though the compression ratio is also higher. In addition, unlike to~\cite{yong2015compression} which does not seem to work well when the size of the convolutional kernel is compressed from \(3 \times 3\) to \(2 \times 2\), our method is able to compress that dimension too while maintaining similar level of accuracy. Finally, our method outperforms MobileNet~\cite{mobilenets} by a large margin.

In the same table, we also report the performance of a variant of our method, using an MPS decomposition on the weights rather than a Tucker one. This result shows that our method works effectively with other decomposition methods as well. Nevertheless, we focused mainly on Tucker as it is the most flexible compression method, allowing us to control the rank of each mode of the weight tensor.

\textbf{Results on face segmentation.}
Finally, we selected two of our best performing models and retrained them for the task of semantic facial part segmentation. Our method offers significant compression ratios (up to 7x) with virtually no loss in accuracy (see Table~\ref{table:results-segmentation}). These results further confirm that our method is task-independent.

 \section{Conclusions}
We proposed an end-to-end trainable method to jointly capture the full structure of a fully-convolutional neural network, by parametrizing it with a single, high-order low-rank tensor. The modes of this tensor represent each of the architectural design parameters of the network (e.g. number of convolutional blocks, depth, number of stacks, input features, etc). This parametrization allows for a joint regularization of the whole network. The number of parameters can be drastically reduced by imposing a low-rank structure on the parameter tensor. We show that our approach can achieve superior performance with low compression rates, and attain high compression rates with negligible drop in accuracy, on both the challenging task of human pose estimation and semantic face segmentation. 
{\small
\bibliographystyle{ieee_fullname}

}

\end{document}